\documentclass[letterpaper, 10 pt, conference]{ieeeconf}  

\IEEEoverridecommandlockouts                              

\title{\LARGE \bf
DXQ-Net: Differentiable LiDAR-Camera Extrinsic Calibration Using Quality-aware Flow
}

\author{Xin Jing$^{1,2}$, Xiaqing Ding$^{2}$, Rong Xiong$^{1}$, Huanjun Deng$^{2}$, Yue Wang$^{1}$
\thanks{*This work was supported by the National Key R\&D Program of China under Grant 2018YFB1600804 and ZheJiang Program in Innovation, Entrepreneurship and Leadership Team (2018R01017).}
\thanks{*This work was supported by Alibaba Group through Alibaba Innovative Research Program.}
\thanks{$^{1}$State Key Laboratory of Industrial Control and Technology, Zhejiang University, Hangzhou, P.R. China.}
\thanks{$^{2}$Alibaba Group, Hangzhou, 310052, China.}
\thanks{*Corresponding author Yue Wang {\tt\small wangyue@iipc.zju.edu.cn}}
}
\usepackage{amsmath}
\usepackage{booktabs}
\usepackage{multirow}
\usepackage{graphicx}
\usepackage{amsfonts,amssymb}
\begin{document}
\maketitle
\thispagestyle{empty}
\pagestyle{empty}
\begin{abstract}
Accurate LiDAR-camera extrinsic calibration is a precondition for many multi-sensor systems in mobile robots. Most calibration methods rely on laborious manual operations and calibration targets. While working online, the calibration methods should be able to extract information from the environment to construct the cross-modal data association. Convolutional neural networks (CNNs) have powerful feature extraction ability and have been used for calibration. However, most of the past methods solve the extrinsic as a regression task, without considering the geometric constraints involved. In this paper, we propose a novel end-to-end extrinsic calibration method named DXQ-Net, using a differentiable pose estimation module for generalization. We formulate a probabilistic model for LiDAR-camera calibration flow, yielding a prediction of uncertainty to measure the quality of LiDAR-camera data association. Testing experiments illustrate that our method achieves a competitive with other methods for the translation component and state-of-the-art performance for the rotation component. Generalization experiments illustrate that the generalization performance of our method is significantly better than other deep learning-based methods.

\end{abstract}

\section{Introduction}

LiDAR-camera fusion is important for mobile robots, benefiting various tasks ranging from semantic understanding \cite{r1} to geometric reconstruction \cite{r2}. The prerequisite enabling the fusion is the accurate extrinsic calibration between the two sensors. However, due to the collision and vibration in practice, the extrinsic may drift in long term, which is harmful to all downstream fusion components. Therefore, an online extrinsic calibration method is desired.

Previous works on online extrinsic calibration make use of the consistency between LiDAR and camera trajectory. With the tracking of visual and LiDAR features, the extrinsic, which is modeled in the state, is estimated online by nonlinear optimization or filtering \cite{r3,r4}. The convergence of these methods is guaranteed by the theory. But these methods have degenerate motion patterns, during which the extrinsic is unobservable. For example, a typical degenerate motion pattern is straight moving on the planar ground, which, unfortunately, is common for mobile robots \cite{r5}. 

Note that the high accuracy of offline extrinsic calibration methods mainly depends on the calibration target, which provides cross-modal feature associations between LiDAR and vision observations \cite{r6}. Such information allows the extrinsic calibration even when the robot is static. Following this idea, deep learning methods are employed to express the correlation between the RGB image and the depth image projected from the point cloud without depending on the calibration target. Based on the correlation volume, the extrinsic is predicted using a fully connected neural network (FC) as a regression task \cite{r7}. However, such architecture may cause performance deterioration when generalizing to unseen scenes or LiDAR-camera configurations. In later works \cite{r8}, calibration flow is proposed, and traditional Perspective-N-Point (PnP) is applied as a post-processing step to solve the extrinsic from the LiDAR-camera flow, increasing the generalization performance. To suppress the flow outliers, these methods additionally apply random sample consensus (RANSAC) for explicit correspondence selection with manually tuned coefficients. These extra steps are non-differentiable, so only the ground truth of the flow is used as supervision, thus it is hard to bring the self-awareness of the flow prediction quality.

\begin{figure*}
\centering
\includegraphics[width=16cm]{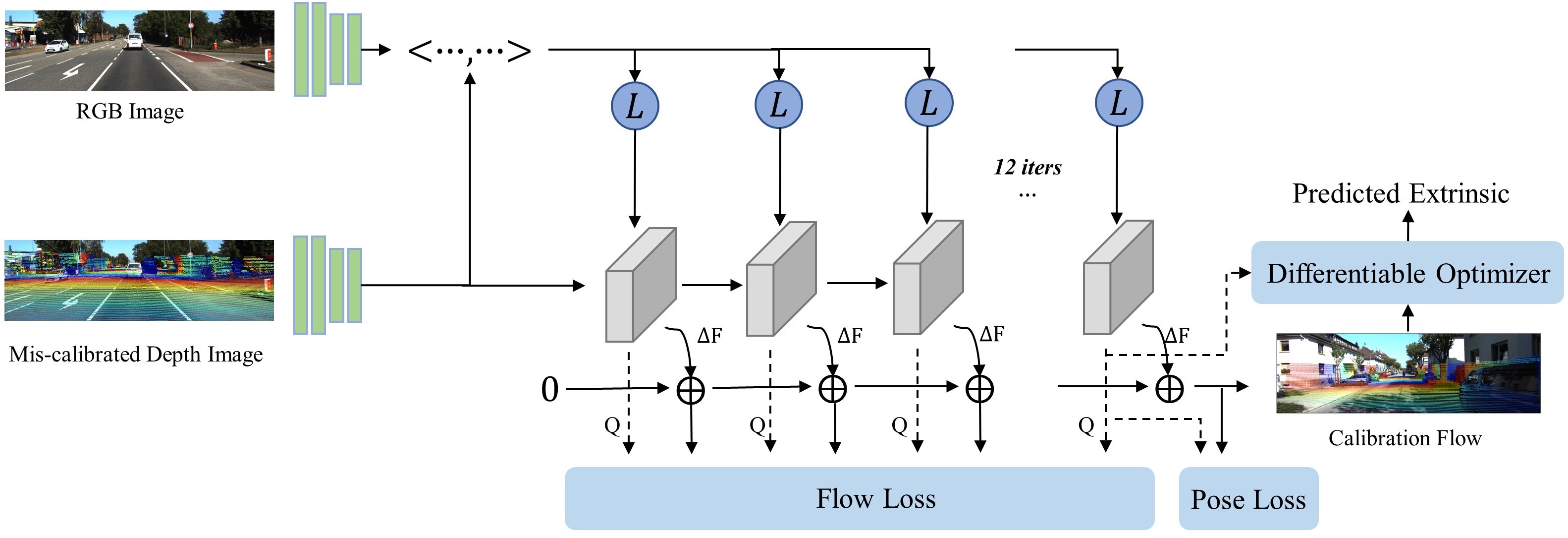}
\caption{Overview of our method. The input of the network is RGB image and miscalibrated depth image. The calibration flow predicted by the network is used to correct the initial projected 2D coordinates. Moreover, the uncertainty of the calibration flow is estimated for each point by the network. Then the 2D-3D correspondences with uncertainty between camera and LiDAR are detected. Finally, the extrinsic parameters are estimated by a gated optimizer.}
\label{fig3}
\end{figure*}

In this paper, we present an end-to-end LiDAR-camera extrinsic calibration network, named DXQ-Net, with a differentiable pose estimation module, which can back-propagate the error from final extrinsic to the flow prediction network. Compared to the methods using FC for direct regression, the learning-free differentiable pose estimation brings inductive bias by the architecture, resulting in better generalization performance. In addition, we further state the network in a probabilistic way, so that the uncertainty of flow is considered in the pose estimation to weaken the outliers in the extrinsic estimation. This uncertainty also acts as a measure of quality, rendering a quantitative selection of the scene for calibration. In the experiments, the results validate the accuracy and generalization of the DXQ-Net, and demonstrate a significant correlation between the quality and the extrinsic error. The architecture of our method is shown in Figure. \ref{fig3}. In summary, the contribution is three folded:
\begin{itemize}
    \item An end-to-end lidar-camera extrinsic calibration network, DXQ-Net, using a differentiable pose estimation module for generalization.
    \item A probabilistic modeling of flow network, yielding a prediction of uncertainty to measure the quality of each lidar-camera data association.
    \item Testing and generalization experiments to verify the performance of DXQ-Net, as well as the correlation between quality and calibration error.
\end{itemize}

\section{Related Work}
In this paper, we focus on the extrinsic calibration between LiDAR and camera, i.e., estimating the rigid transformation matrix between these two sensors. According to whether specific calibration targets are used, the LiDAR-camera extrinsic calibration methods can be categorized into target-based and targetless. Moreover, due to the powerful feature extraction capability of convolutional neural networks (CNNs), deep learning-based calibration methods have been proposed in recent years.

\subsection{Target-based Methods}
Geometric solids \cite{r9} and chessboards \cite{r10} have been widely applied in target-based methods due to the convenience of building cross-modal data association. The 3D structure of chessboards is recovered by extracting corner points from images in the camera coordinate. The constraint that the chessboard planes should coincide with the LiDAR planes is used for optimization to solve the extrinsic parameters \cite{r11}. The above method requires that LiDAR and camera must have a common field of view. To eliminate the constraint, Bo. \cite{r6} proposes to move the camera to reconstruct the 3D environment using the sequential images, then align 3D visual points to LiDAR point cloud based on a graph optimization method to calculate extrinsic parameters. There are other different calibration targets, such as monochromatic board \cite{r12} or ArUco markers \cite{r13}, which are used to further enhance the accuracy of the calibration.
\subsection{Targetless Methods}
During the operation of self-driving vehicles, the extrinsic parameters will inevitably be biased. Therefore, vehicles should have the ability to detect deviations and re-calibrate in natural environments. Online targetless calibration algorithms without specific calibration targets can deal with this problem.

Levinson et al. \cite{r14} propose that if the extrinsic parameters are correct, the depth discontinuity points detected in the point cloud will be more likely to coincide with the edges on the image when projected onto the image. Extrinsic calibration is modeled as an alignment optimization problem between depth discontinuity edges of the point cloud and image edges. Another paper \cite{r15} improves the above method. It proves that depth-continuous edges are more effective for calibration, and proposes a 3-D edge extraction algorithm based on plane fitting. Taylor et al. \cite{r16} proposes a new measurement parameter, gradient orientation measure, to describe the gradient correlation of LiDAR point clouds and images. Ishikawa et al. \cite{r3} converts the LiDAR-camera calibration problem to a hand-eye calibration problem. This method can provide the initial value of extrinsic parameters, but it depends heavily on the accuracy of the odometry \cite{r17}.
\subsection{Deep Learning Methods}
Deep neural networks have demonstrated powerful feature extraction capabilities in several fields like object detection and semantic segmentation. To our best knowledge, RegNet \cite{r18} uses a convolutional neural network(CNN) for feature extraction and matching, then uses MLP network regression to get the six DOF of extrinsic parameters. In RegNet, the Euclidean loss function is used directly and the geometric constraints of extrinsic parameters are ignored. CalibNet \cite{r19} uses a 3D spatial transformer layer (3D STL) to deal with this problem. The network is trained by maximizing the geometric and photometric consistency between the color image and the point cloud. RGGNet \cite{r20} considers Riemannian geometry constraints in loss function. The deep generative model is used to build a tolerance-aware loss function. LCCNet \cite{r7} exploits the cost volume layer to learn the correlation between the color image and the depth image generated by the point cloud. CFNet \cite{r8} defines calibration flow to illustrate the deviation of the initial projection from the ground truth. Encoder-decoder network is used to predict calibration flow. Then The EPnP-RANSAC algorithm is applied to estimate the extrinsic parameters with 2D–3D correspondences constructed by the calibration flow. 

\section{Method}
Figure. \ref{fig3} provides an overview of our method. The network takes in a pair of the color images and miscalibrated depth images, $(I, D)$, and outputs the calibration flow with uncertainty, which provides the 2D-3D correspondences with uncertainty between LiDAR and camera. The differentiable pose estimator module is then applied to estimate the extrinsic parameters and constrains the flow with uncertainty during training.

\subsection{Calibration Flow}
We use the pinhole projection model and assume that the camera intrinsic is known. Given ground truth extrinsic $T_{gt}$ and camera intrinsic $K$, we can transform each 3D LiDAR point $P_i = [X_i, Y_i, Z_i]\in \mathbb{R}^3$ into the camera coordinate then project it onto the image plane. We denote the projected 2D coordinates as $p_i = [u_i, v_i] \in \mathbb{R}^2$ and the inverse depth as $d_i = 1 / Z_i$. The projection process is expressed as follows:
\begin{equation}
    [\hat{X}_i, \hat{Y}_i, \hat{Z}_i, 1]^T= T_{gt} [X_i, Y_i, Z_i, 1]^T
\end{equation}

\begin{equation}
\begin{bmatrix}
    u_i\\ 
    v_i\\ 
    d_i
    \end{bmatrix}
    = 
    \begin{bmatrix}
        f_x\cdot(\hat{X}_i/\hat{Z}_i) + c_x \\
        f_y\cdot (\hat{Y}_i/\hat{Z}_i) + c_y \\
        1/\hat{Z}_i
    \end{bmatrix}
\end{equation}
where $(f_x, f_y, c_x, c_y)$ are the camera intrinsic parameters. The depth image can then be generated by preserving the largest inverse depth on each projected image pixel.

During online operation, the extrinsic parameters would drift and lead to a deviation between the initial extrinsic parameter $T_{init}$ and the ground truth $T_{gt}$. 
Therefore, the coordinate of the mis-calibrated projection point $\widetilde{p}_i$ projected by $T_{init}$ is different from $p_i$. We denote the deviation between $\widetilde{p}_i$ and $p_i$ as the calibration flow $F$. 
The calibration flow is similar to the optical flow, which includes two channels and represents the deviations in the horizontal and vertical directions, respectively. The ground truth calibration flow $F_{gt}$ is calculated as follows:
\begin{equation}
    F_{gt}(\widetilde{p}_{i}) = [u_i - \widetilde{u}_i, v_i - \widetilde{v}_i]
\end{equation}
\subsection{Network Architecture}
Our network architecture is derived from RAFT \cite{r21} and the details are shown in Figure. \ref{fig1}. We construct a 4D correlation volume by computing feature similarity between all pairs of pixels of the color image and the miscalibrated depth image, then estimate the flow iteratively with uncertainty using a recurrent GRU-unit. Finally, the extrinsic parameters are optimized based on the calibration flow and uncertainty. 
\subsubsection{Feature Correlation Pyramid}
We first extract features from the input color image and miscalibrated depth image. Two separate feature encoders $f_{\theta_1}$ and $f_{\theta_2}$ are utilized, which share the same structure but different weights as the inputs have different modalities. 
Each feature encoder consists of 6 residual blocks and the resolution is downsampled 1/2 every 2 blocks. The output feature maps are 128-dimension with 1/8 resolution compared with the input images.

We construct a 4D correlation volume by computing the dot product between all pairs of feature maps between the input color image and the miscalibrated depth image.
\begin{equation}
    C_{ijkh}(I_1, D_1) = <f_{\theta_1}(I_1)_{ij}, f_{\theta_2}(D_1)_{kh}> \in \mathbb{R}^{H \times W \times H \times W}
\end{equation}
The correlation volume is then pooled with three $2 \times 2$ average pooling kernels along the last two dimensions. We then integrate the output of each pooling operation into a correlation pyramid $\{C_1, C_2, C_3, C_4\}$ 
\begin{equation}
    C_k \in \mathbb{R} ^ {H \times W \times H / 2^{k-1} \times W / 2^{k-1}}, k=1,2,3,4
\end{equation}

\subsubsection{Probabilistic Flow Estimator}
For each depth image pixel, we can map it to its RGB image correspondence based on the calibration flow and index from the correlation pyramid to produce a set of correlation features within a certain radius. The indexing operation can be referred to RAFT \cite{r21}. Then the retrieved features and the calibration flow with uncertainty are put into the probabilistic flow estimator for flow update and uncertainty estimation. 
The probabilistic flow estimator is a recurrent GRU-unit $f_r$ which outputs a delta flow $\Delta F$ and the updated uncertainty $Q\in \mathbb{R}^{W/8\times H/8}$ during each iteration.
Different from RAFT which uses a series of 1x5 and 5x1 GRU units, we use a single 3x3 unit instead. 
The calibration flow and the corresponding uncertainty are initialized as zero before the first iteration.

Referring to probabilistic regression techniques for optical flow \cite{r22} and other tasks \cite{r23}, we formulate the estimation of calibration flow as a probabilistic model, which makes the network learn both the flow and its uncertainty. Therefore, our goal is to learn the conditional probability density $p(\Delta F|I, D;f_{\theta_1}, f_{\theta_2}, f_r)$ of the delta flow $\Delta F$ given the input RGB and depth images. It can be achieved by making the network predict the variance of the estimate $\Delta F$. In our cases, the predicted density $p(\Delta F|Q)$ is modeled as Gaussian distribution. The density is given by,
\begin{equation}
    p(\Delta F|\mu, \sigma^2) = \frac{1}{\sqrt{2\pi}\sigma_u} e^{-\frac{(\Delta F_u-\mu_u)^2}{2{\sigma^2_u}}}\cdot\frac{1}{\sqrt{2\pi}\sigma_v} e^{-\frac{(\Delta F_v-\mu_v)^2}{2{\sigma^2_v}}}
\end{equation}
where $\Delta F=[\Delta F_u, \Delta F_v]$, the mean $\mu=[\mu_u, \mu_v]$ and the variance $\sigma=[\sigma_u, \sigma_v]$. We model the two components $u$ and $v$ of the flow vector as conditionally independent.

After each iteration $t$, the calibration flow $F_{t}$ will be updated based on the output $\Delta F_t$ from this iteration and the last updated calibration flow $F_{t-1}$ 
\begin{equation}
F_{t} = F_{t-1} + \Delta F_t, t=1,2,\dots,N
\end{equation}
where $N$ is the total number of iterations. During experiments, we set $N$ as $12$.

\subsubsection{Differentiable Pose Estimation Module}
To back propagate the extrinsic error to the flow prediction network, the differentiable pose estimation module is designed for solving extrinsic parameters. It takes the predicted flow $F_N = [F_{N_{u}}, F_{N_{v}}]$ and uncertainty $Q_N$ as input. For any 3D LiDAR point $P_i$, we first update its originally projected 2D image point $\widetilde{p}_i$ into $\hat{p}_i$ with the predicted flow
\begin{equation}
\hat{p}_i=\widetilde{p}_i+F_N(\widetilde{p}_i)
\end{equation}

Then the initial extrinsic parameters $T_{init}$ can be optimized by minimizing the reprojection errors
\begin{equation}
    \Delta T^* = arg\mathop{min}\limits_{\Delta T} \frac{1}{2} \sum_{\widetilde{p}_i\in S_v} \Vert \hat{p}_i - K T_{init}\Delta TP_i \Vert_{Q_N(\widetilde{p}_i)}
\end{equation}
where $\Delta T \in \mathbb{SE}(3)$ and $S_v$ denotes the set of projected points that are within the RGB image scope. We solve this optimization problem by Gauss-Newton method \cite{r24}.

The optimization algorithm presented here is end-to-end differentiable, based on which the extrinsic error can be back-propagated to optimize the flow and uncertainty estimation, making the quality of the flow uncertainty consistent with the calibration estimation.

During inference, we take the normalized uncertainty as a measure of flow quality and utilize it to filter out points with high uncertainty.
The rest of the points form a high-quality point set $S_q$. The threshold was set to 0.5 in our experiments. Therefore, the pose estimation module works as a gated optimizer during inference. The optimization will be formulated as:
\begin{equation}
    \Delta T^* = arg\mathop{min}\limits_{\Delta T} \frac{1}{2} \sum_{\hat{p}_i \in S_q} \Vert \hat{p}_i - K T_{init} \Delta TP_i \Vert_2^2
\end{equation}

\subsection{Loss Function}
Our network is supervised based on the flow loss $L_f$ and geodesic loss $L_g$. 
\subsubsection{Flow Loss}
The flow loss is defined as the $L1$ distance between the predicted and ground truth calibration flow on valid pixels over the full sequence of predictions, ${F_1, F_2, ..., F_N}$, with exponentially increasing weights. The flow is calculated as follows:
\begin{equation}
    L_f = \sum_{i=1}^{N} \gamma^{N-i} (Q_i \cdot \Vert F_{gt} - F_i \Vert_1 + 1 / Q_i)
\end{equation}
where $\gamma$ is set to 0.8 in our experiments.
\subsubsection{Geodesic Loss}
After obtaining the predicted extrinsic parameters by the differentiable pose estimation module, the geodesic loss can be defined as:
\begin{equation}
    L_g = \Vert Log((T_{init}\Delta T)^{-1} \cdot T_{gt}) \Vert_1
\end{equation}
where the Log is the logarithm map from the manifold to the Euclidean space.

Our final loss function consists of a weighted sum of the flow loss and geodesic loss:
\begin{equation}
    L_{total} = \lambda_f L_f + \lambda_g L_g
\end{equation}
where $\lambda_f$ is the weight of the flow loss, $\lambda_g$ is the weight of the geodesic loss.

\section{Experiments}
We evaluate our method on KITTI-odometry \cite{r25}, KITTI-detection \cite{r25} and KITTI-360 \cite{r26} datasets, in which synchronized Velodyne point clouds and RGB images are collected from different scenes. The input point cloud and image pairs are aligned according to the timestamps.

\subsection{Data Preparation}
As with many previous methods \cite{r7, r8}, we expand the training data by applying random perturbations to the ground truth extrinsic parameters $T_{gt}$ in the datasets. Considering that during online localization, the extrinsic drift slowly and would not be too far away from the ground truth, we set the random translational perturbations within $(\pm10cm)$ and random rotational perturbations within $(\pm5^\circ)$. 
During each training iteration, we calculate the perturbed transformation matrix $\Delta T$ by the translation vector $\Delta t$ and the rotation vector $\Delta r$, which are chosen randomly within the initial error range. The perturbed rotation vector is in the form of Euler angles. By adding $\Delta T$ to $T_{gt}$, the initial extrinsic $T_{init} = T_{gt}\cdot (\Delta T)^{-1}$ is obtained. The miscalibrated depth image is generated by projecting the LiDAR point cloud on the image plane with the initial extrinsic matrix $T_{init}$ and camera intrinsic matrix $K$. The network takes the miscalibrated depth image and the color image as input.
\subsubsection{KITTI odometry dataset}
The dataset consists of 21 sequences from different scenes. Like LCCNet \cite{r7}, we used sequences from 00 to 20 for training and validation (39011 frames) and sequence 00 for testing (4541 frames). Specifically, only the LiDAR point cloud and the left color image are used for training. To test the precision of the proposed method and the generalization ability across different cameras, we evaluate the calibration on both the left and right color images in the experiments.
\subsubsection{KITTI detection dataset}
The dataset consists of an official training set (7481 frames) and a testing set (7518 frames). The camera is the same with the one used in the KITTI odometry dataset, while the scenarios are much different. Our model is evaluated on the training set to test the generalization performance when the scenario changes.
\subsubsection{KITTI-360 dataset}
Compared to the previous two datasets, the acquisition scenarios and the cameras used in the KITTI-360 dataset are both different. We fine-tune the pre-trained model with 4000 frames from sequence 00 in the KITTI-360 dataset. Other sequences were selected as test sequences. For comparison, we also evaluate the pre-trained model on the KITTI odometry dataset directly on the test sequences of the KITTI-360 dataset.

\begin{table*}
    \caption{The calibration results on the KITTI odometry dataset.}
    \centering
    \begin{tabular}{*{10}{c}}
    \toprule
    \multirow{2}*{Initial error range} & & \multicolumn{4}{c}{Translation(cm)} & \multicolumn{4}{c}{Rotation($^\circ$)}\\
    & & $E_t$ & $E_x$ & $E_y$ & $E_z$ & $E_r$ & $E_{roll}$ & $E_{pitch}$ & $E_{yaw}$\\
    \midrule
    \multirow{3}*{($\pm10cm$, $\pm5^\circ$)} & Mean & 1.425 & 0.754 & 0.476 & 1.091 & 0.084 & 0.049 & 0.046 & 0.032\\
    & Median & 0.807 & 0.529 & 0.512 & 0.783 & 0.069 & 0.032 & 0.033 & 0.022\\
    & Std & 1.011 & 1.044 & 0.824 & 0.982 & 0.096 & 0.078 & 0.048 & 0.053\\
    \bottomrule
    \label{tab1}
    \end{tabular}
\end{table*}

\subsection{Evaluation Metrics}
The experimental results are analyzed according to the translation and rotation errors of the predicted extrinsic parameters $T_{pred}$. The translation error is evaluated by the 2-norm of the difference of translation vectors. The translation error is expressed as follows:
\begin{equation}
    \delta T=(T_{pred})^{-1}\cdot T_{gt}=
    \begin{bmatrix}
    \delta R & \delta t \\
    0_{1\times3} & 1
    \end{bmatrix}
\end{equation}
\begin{equation}
    E_t = \Vert \delta t\Vert_2
\end{equation}
The translation errors along three axes $E_x, E_y, E_z$ are also evaluated. 
The angular error $E_r$ is defined as the 2-norm of the Euler angle vector $[E_{roll}, E_{pitch}, E_{yaw}]^T$ derived from the rotation error matrix $\delta R$
\begin{equation}
    \delta R = 
    \begin{bmatrix}
        r_{11} & r_{12} & r_{13} \\
        r_{21} & r_{22} & r_{23} \\
        r_{31} & r_{32} & r_{33}
    \end{bmatrix}
\end{equation}
\begin{equation}
    \begin{aligned}
        E_{roll} &= atan2(r_{32}, r_{33}) \\
        E_{pitch} &= atan2(-r_{31}, \sqrt{r^2_{31}+r^2_{33}})\\
        E_{yaw} &= atan2(r_{21}, r_{11}) 
    \end{aligned}
\end{equation}

\subsection{Training Details}
During the training stage, AdamW \cite{r27} Optimizer is used with an initial learning rate 3e$-$5. The weight decay is set as 4e$-$4. The 1Cycle \cite{r28} learning rate scheduler is used during training. Our proposed calibration network is trained on four Nvidia 2080Ti GPU with batch size 32 and total epochs 80. The network weights were initialized using kaiming initialization \cite{r29}. When fine-tuning the pre-trained model on the KITTI360 dataset, we both evaluate the model trained after 1 and 10 epochs.

\begin{table}
    \caption{The comparison results on the KITTI odometry dataset.}
    \centering
    \begin{tabular}{*{7}{c}}
    \toprule
     & \multicolumn{3}{c}{Translation Error(cm)} & \multicolumn{3}{c}{Rotation Error($^\circ$)}\\
     & Mean & Median & Std & Mean & Median & Std\\
    \midrule
    CFNet & 1.724 & 1.692 & \textbf{0.758} & 0.171 & 0.147 & 0.101\\
    SemAlign & / & / & / & 1.14 & 0.46 & /\\
    LCCNet & \textbf{1.292} & \textbf{0.612} & 1.959 & 0.173 & 0.116 & 0.456\\
    Ours & 1.425 & 0.808 & 1.011 & \textbf{0.084} & \textbf{0.069} & \textbf{0.096}\\
    \bottomrule
    \label{tab2}
    \end{tabular}
\end{table}

\begin{table}
    \caption{Ablation study}
    \centering
    \setlength\tabcolsep{2pt}
    \begin{tabular}{*{8}{c}}
    \toprule
    \multirow{2}*{Exp} & & \multicolumn{3}{c}{Translation Error(cm)} & \multicolumn{3}{c}{Rotation Error($^\circ$)}\\
    &  & Mean & Median & Std & Mean & Median & Std\\
    \midrule
    \multirow{4}*{Exp.0} & base(flow loss) & 1.704 & 1.241 & 1.184 & 0.102 & 0.074 & 0.243\\
    & + differentiable loss & 1.592 & 0.986 & 1.321 & 0.092 & 0.081 & 0.177 \\
    & + quality aware & 1.425 & 0.808 & \textbf{1.011} & \textbf{0.084} & \textbf{0.069} & \textbf{0.096}\\
    & LCCNet & \textbf{1.292} & \textbf{0.612} & 1.959 & 0.173 & 0.116 & 0.456\\
    \midrule
    \multirow{4}*{Exp.1} & base(flow loss) & 3.361 & 2.422 & 1.943 & 0.172 & 0.132 & 0.562\\
    & + differentiable loss & 3.028 & 2.341 & 1.802 & 0.164 & 0.127 & 0.344\\
    & + quality aware & \textbf{2.943} & \textbf{2.278} & \textbf{1.791} &\textbf{0.155} & \textbf{0.125} & \textbf{0.101}\\
    & LCCNet & 52.489 & 52.494 & 0.251 & 1.541 & 1.467 & 0.574\\
    \midrule
    \multirow{4}*{Exp.2} & base(flow loss) & 3.124 & 2.499 & 1.745 & 0.173 & 0.144 & 0.557\\
    & + differentiable loss & 2.506 & 2.473 & 1.749 & 0.145 & \textbf{0.116} & 0.324\\
    & + quality aware & \textbf{2.223} & \textbf{2.462} & \textbf{1.414} & \textbf{0.140} & 0.123 & \textbf{0.084}\\
    & LCCNet & 2.631 & 2.487 & 1.372 & 0.398 & 0.178 & 0.895\\
    \midrule
    \multirow{6}*{Exp.3} & base(flow loss) & 6.942 & 5.624 & 2.343 & 3.443 & 1.672 & 2.913\\
    & + differentiable loss & 6.163 & 4.779 & 2.012 & 3.401 & 2.791 & 1.469\\
    & + quality aware & 5.648 & 4.697 & 1.264 & 2.889 & 1.028 & 1.336\\
    & + 1 epoch & 4.323 & 3.125 & 1.165 & 1.928 & 1.103 & 1.029\\
    & + 10 epoch & \textbf{2.986} & \textbf{1.834} & \textbf{1.065} & \textbf{0.689} & \textbf{0.414} & 0.936\\
    & LCCNet & 79.614 & 62.458 & 1.894 & 3.045 & 1.658 & \textbf{0.568}\\
    \bottomrule
    \label{tab3}
    \end{tabular}
\end{table}

\subsection{Results and Discussion}
\subsubsection{Results on KITTI-Odometry}
The detailed calibration results on the KITTI odometry dataset are shown in TABLE \ref{tab1}. Our approach achieves a mean translation error of 1.425cm (x, y, z: 0.754cm, 0.476cm, 1.091cm), and a mean rotation error of $0.084^\circ$ (roll, pitch, yaw: $0.049^\circ$ , $0.046^\circ$ , $0.032^\circ$).

We compare results on the KITTI odometry dataset with the other methods and the results are shown in TABLE \ref{tab2}. 
It can be seen that our method achieves the best rotational estimation performance and competitive translational estimation performance even compared with the regression-based methods like LCCNet. 
 
\begin{figure*}
\centering
\includegraphics[width=17cm]{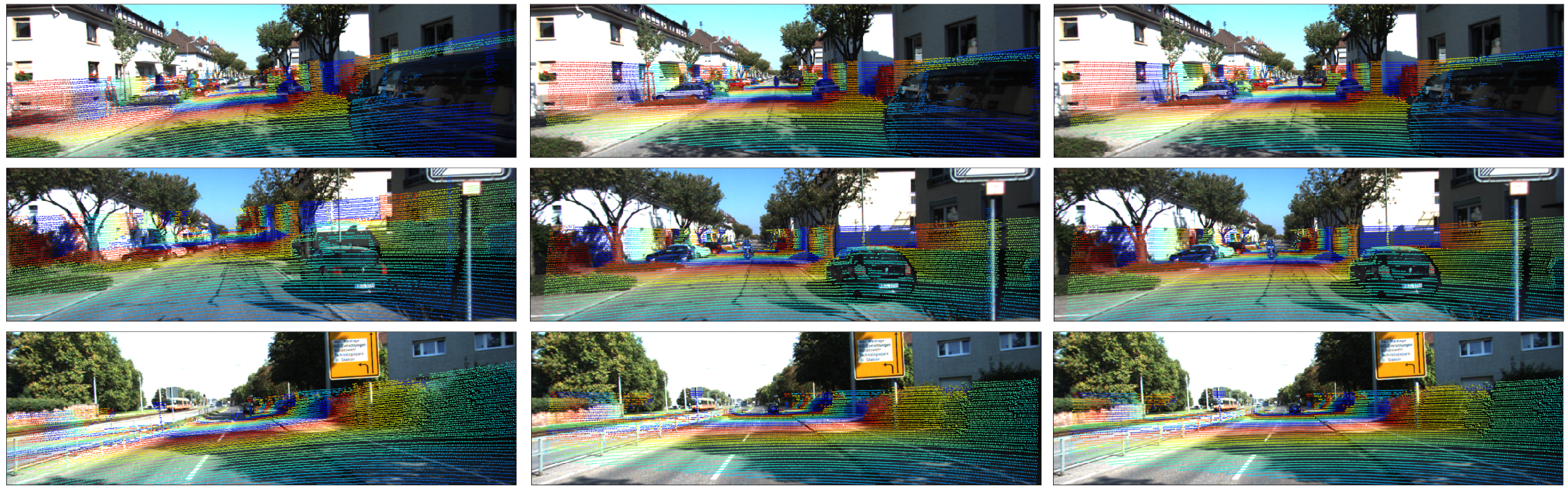}
\caption{Examples of the projection results obtained from the predicted extrinsic parameters. Each row of images from top to bottom is the results on KITTI odometry dataset(left color camera), KITTI odometry dataset(right color camera) and KITTI detection dataset. Each column of images from left to right indicates the mis-calibrated depth images, the predicted projection results and the ground truth projection depth images.}
\label{fig1}
\end{figure*}

Figure. \ref{fig1} shows some examples of the predictions of our method in different working scenarios. It can be seen that the projected depth image generated by the predicted extrinsic parameters is almost the same as the ground truth. 

\subsubsection{Ablation Study}

Three different models are trained to verify the validity of the probabilistic modeling and differentiable pose estimation module in terms of precision and generalization. The first model is the baseline, which is set up in the same way as CFNet \cite{r8}. Only the flow loss is used for training. The second model adds a differentiable pose estimation module and geodesic loss based on the first model. The third model is our complete model with probabilistic modeling. 
 
We choose sequence 00 of the KITTI odometry dataset for evaluation, which is marked as Exp.0. Moreover, to validate the contribution of our design to the generalization performance of the model, we perform three extra ablation experiments in generalization scenarios, marking them as Exp.1, Exp.2, Exp.3. 
The first experiment Exp.1 is designed to test the generalization performance across different cameras.
The LiDAR point clouds and right color images in sequence 00 of the KITTI odometry dataset are used for evaluation. 
The second experiment Exp.2 is designed to test the generalization performance across different scenarios. 
The training set of the KITTI detection dataset is used for evaluation. The third experiment Exp.3 is designed to test the generalization performance when both the cameras and scenarios are changed. 
The KITTI-360 dataset sequence 0001 is used for evaluation. Since the camera parameters in KITTI-360 dataset are very different compared to the training set, we further fine-tune the model using the 0000 sequence in the KITTI-360 dataset after validating the pre-trained model. The results of our ablation study are shown in the TABLE. \ref{tab3}. As can be seen, the differentiable pose estimator module and quality-aware probabilistic modeling design contribute to the performance and generalization capability of our method.
 
\begin{figure}
\centering
\includegraphics[width=7cm]{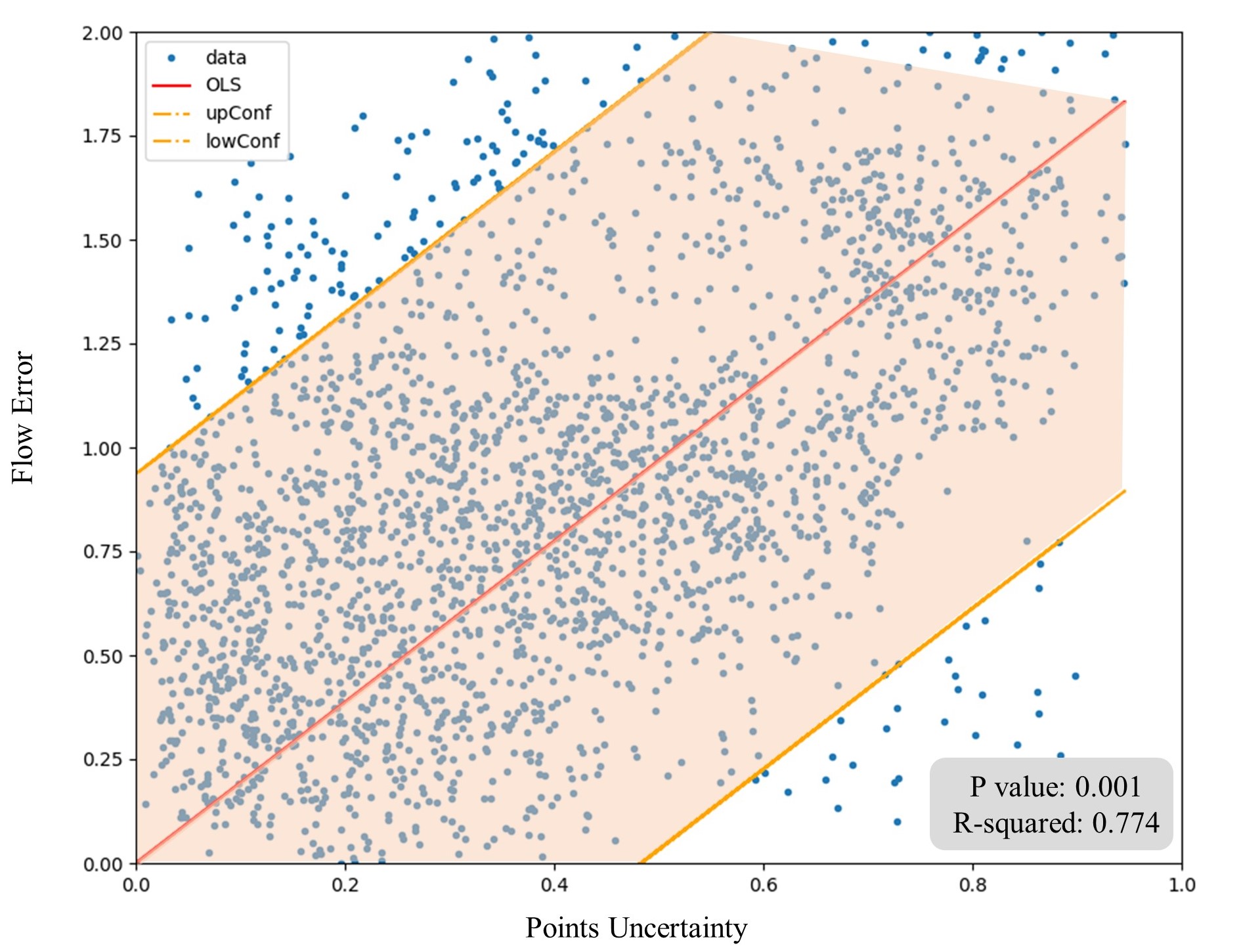}
\caption{Linear regression model of flow error and points uncertainty Q.}
\label{fig2}
\end{figure}

To further verify the quality of flow uncertainty output from the network, the relationship between uncertainty and flow error is analyzed. The flow error of point $\widetilde{p}_i$ is the error between predicted flow and ground truth flow, which is defined as follows:
\begin{equation}
   E_f = \Vert F_{N}(\widetilde{p}_i) - F_{gt}(\widetilde{p}_i) \Vert_2
\end{equation}
The relationship is shown in Figure. \ref{fig2}. A linear regression model is used to fit the relationship between the two variables. The orange area represents the 95 percent confidence interval. It can be seen that most points are within the 95 percent confidence interval. The R-squared value of the model is 0.774. The p-value is 0.001. Therefore, the flow error is positively correlated with points uncertainty significantly. It demonstrates the effectiveness of the uncertainty estimation from the network.

\subsubsection{Generalization Comparison Experiments}
We also compare our method with the other state-of-the-art deep learning-based calibration methods LCCNet and CFNet under three generalization scenarios Exp.1, Exp.2 and Exp.3. We use the open-source code of LCCNet to obtain the results of LCCNet under the generalization scenarios. There is no open-source code of CFNet. In the above ablation experiments, the baseline model is almost identical to CFNet, and obtain almost the same results as CFNet on the validation set of the KITTI odometry dataset, so it is feasible to use our baseline model as the substitute for comparison. TABLE \ref{tab3} shows the comparison results. It is obvious that our method has better generalization performance than LCCNet and CFNet.

\section{CONCLUSIONS}

In this paper, we introduce a novel method for extrinsic calibration between LiDAR and camera. We innovatively formulate the estimation of calibration flow as a probabilistic model, which makes the network predict flow as well as its uncertainty. A differentiable pose estimation module is further applied to back-propagate the extrinsic error for end-to-end network training. In the test dataset, our model yields a mean translation error $1.425cm$ and $0.084^\circ$, which is better than the state-of-the-art methods. Generalization experiments prove that our model can generalize to unseen scenes and different cameras and have superior performance compared with the other deep learning-based methods.





\section*{ACKNOWLEDGMENT}

This work was supported by Alibaba Group through Alibaba Innovative Research Program.

This work was supported by the National Key R\&D Program of China under Grant 2018YFB1600804 and ZheJiang Program in Innovation, Entrepreneurship and Leadership Team (2018R01017).

\bibliographystyle{IEEEtran}
\bibliography{reference}

\begin{thebibliography}{10}
\providecommand{\url}[1]{#1}
\csname url@samestyle\endcsname
\providecommand{\newblock}{\relax}
\providecommand{\bibinfo}[2]{#2}
\providecommand{\BIBentrySTDinterwordspacing}{\spaceskip=0pt\relax}
\providecommand{\BIBentryALTinterwordstretchfactor}{4}
\providecommand{\BIBentryALTinterwordspacing}{\spaceskip=\fontdimen2\font plus
\BIBentryALTinterwordstretchfactor\fontdimen3\font minus
  \fontdimen4\font\relax}
\providecommand{\BIBforeignlanguage}[2]{{%
\expandafter\ifx\csname l@#1\endcsname\relax
\typeout{** WARNING: IEEEtran.bst: No hyphenation pattern has been}%
\typeout{** loaded for the language `#1'. Using the pattern for}%
\typeout{** the default language instead.}%
\else
\language=\csname l@#1\endcsname
\fi
#2}}
\providecommand{\BIBdecl}{\relax}
\BIBdecl

\bibitem{r1}
Z.~Zhuang, R.~Li, K.~Jia, Q.~Wang, Y.~Li, and M.~Tan, ``Perception-aware
  multi-sensor fusion for 3d lidar semantic segmentation,'' in
  \emph{Proceedings of the IEEE/CVF International Conference on Computer
  Vision}, 2021, pp. 16\,280--16\,290.

\bibitem{r2}
F.~Nouguier, S.~T. Grilli, and C.-A. Gu{\'e}rin, ``Nonlinear ocean wave
  reconstruction algorithms based on simulated spatiotemporal data acquired by
  a flash lidar camera,'' \emph{IEEE Transactions on Geoscience and Remote
  Sensing}, vol.~52, no.~3, pp. 1761--1771, 2013.

\bibitem{r3}
R.~Ishikawa, T.~Oishi, and K.~Ikeuchi, ``Lidar and camera calibration using
  motions estimated by sensor fusion odometry,'' in \emph{2018 IEEE/RSJ
  International Conference on Intelligent Robots and Systems (IROS)}.\hskip 1em
  plus 0.5em minus 0.4em\relax IEEE, 2018, pp. 7342--7349.

\bibitem{r4}
M.~{\'A}. Mu{\~n}oz-Ba{\~n}{\'o}n, F.~A. Candelas, and F.~Torres, ``Targetless
  camera-lidar calibration in unstructured environments,'' \emph{IEEE Access},
  vol.~8, pp. 143\,692--143\,705, 2020.

\bibitem{r5}
C.~R. Qi, W.~Liu, C.~Wu, H.~Su, and L.~J. Guibas, ``Frustum pointnets for 3d
  object detection from rgb-d data,'' in \emph{Proceedings of the IEEE
  conference on computer vision and pattern recognition}, 2018, pp. 918--927.

\bibitem{r6}
B.~Fu, Y.~Wang, X.~Ding, Y.~Jiao, L.~Tang, and R.~Xiong, ``Lidar-camera
  calibration under arbitrary configurations: Observability and methods,''
  \emph{IEEE Transactions on Instrumentation and Measurement}, vol.~69, no.~6,
  pp. 3089--3102, 2019.

\bibitem{r7}
X.~Lv, B.~Wang, Z.~Dou, D.~Ye, and S.~Wang, ``Lccnet: Lidar and camera
  self-calibration using cost volume network,'' in \emph{Proceedings of the
  IEEE/CVF Conference on Computer Vision and Pattern Recognition}, 2021, pp.
  2894--2901.

\bibitem{r8}
X.~Lv, S.~Wang, and D.~Ye, ``Cfnet: Lidar-camera registration using calibration
  flow network,'' \emph{Sensors}, vol.~21, no.~23, p. 8112, 2021.

\bibitem{r9}
J.~K{\"u}mmerle and T.~K{\"u}hner, ``Unified intrinsic and extrinsic camera and
  lidar calibration under uncertainties,'' in \emph{2020 IEEE International
  Conference on Robotics and Automation (ICRA)}.\hskip 1em plus 0.5em minus
  0.4em\relax IEEE, 2020, pp. 6028--6034.

\bibitem{r10}
J.~Cui, J.~Niu, Z.~Ouyang, Y.~He, and D.~Liu, ``Acsc: Automatic calibration for
  non-repetitive scanning solid-state lidar and camera systems,'' \emph{arXiv
  preprint arXiv:2011.08516}, 2020.

\bibitem{r11}
A.~Geiger, F.~Moosmann, {\"O}.~Car, and B.~Schuster, ``Automatic camera and
  range sensor calibration using a single shot,'' in \emph{2012 IEEE
  international conference on robotics and automation}.\hskip 1em plus 0.5em
  minus 0.4em\relax IEEE, 2012, pp. 3936--3943.

\bibitem{r12}
Y.~Park, S.~Yun, C.~S. Won, K.~Cho, K.~Um, and S.~Sim, ``Calibration between
  color camera and 3d lidar instruments with a polygonal planar board,''
  \emph{Sensors}, vol.~14, no.~3, pp. 5333--5353, 2014.

\bibitem{r13}
A.~Dhall, K.~Chelani, V.~Radhakrishnan, and K.~M. Krishna, ``Lidar-camera
  calibration using 3d-3d point correspondences,'' \emph{arXiv preprint
  arXiv:1705.09785}, 2017.

\bibitem{r14}
J.~Levinson and S.~Thrun, ``Automatic online calibration of cameras and
  lasers.'' in \emph{Robotics: Science and Systems}, vol.~2.\hskip 1em plus
  0.5em minus 0.4em\relax Citeseer, 2013, p.~7.

\bibitem{r15}
C.~Yuan, X.~Liu, X.~Hong, and F.~Zhang, ``Pixel-level extrinsic self
  calibration of high resolution lidar and camera in targetless environments,''
  \emph{IEEE Robotics and Automation Letters}, vol.~6, no.~4, pp. 7517--7524,
  2021.

\bibitem{r16}
D.~Scaramuzza, A.~Harati, and R.~Siegwart, ``Extrinsic self calibration of a
  camera and a 3d laser range finder from natural scenes,'' in \emph{2007
  IEEE/RSJ International Conference on Intelligent Robots and Systems}.\hskip
  1em plus 0.5em minus 0.4em\relax IEEE, 2007, pp. 4164--4169.

\bibitem{r17}
Z.~Taylor and J.~Nieto, ``Motion-based calibration of multimodal sensor
  extrinsics and timing offset estimation,'' \emph{IEEE Transactions on
  Robotics}, vol.~32, no.~5, pp. 1215--1229, 2016.

\bibitem{r18}
N.~Schneider, F.~Piewak, C.~Stiller, and U.~Franke, ``Regnet: Multimodal sensor
  registration using deep neural networks,'' in \emph{2017 IEEE intelligent
  vehicles symposium (IV)}.\hskip 1em plus 0.5em minus 0.4em\relax IEEE, 2017,
  pp. 1803--1810.

\bibitem{r19}
G.~Iyer, R.~K. Ram, J.~K. Murthy, and K.~M. Krishna, ``Calibnet: Geometrically
  supervised extrinsic calibration using 3d spatial transformer networks,'' in
  \emph{2018 IEEE/RSJ International Conference on Intelligent Robots and
  Systems (IROS)}.\hskip 1em plus 0.5em minus 0.4em\relax IEEE, 2018, pp.
  1110--1117.

\bibitem{r20}
K.~Yuan, Z.~Guo, and Z.~J. Wang, ``Rggnet: Tolerance aware lidar-camera online
  calibration with geometric deep learning and generative model,'' \emph{IEEE
  Robotics and Automation Letters}, vol.~5, no.~4, pp. 6956--6963, 2020.

\bibitem{r21}
Z.~Teed and J.~Deng, ``Raft: Recurrent all-pairs field transforms for optical
  flow,'' in \emph{European conference on computer vision}.\hskip 1em plus
  0.5em minus 0.4em\relax Springer, 2020, pp. 402--419.

\bibitem{r22}
J.~Gast and S.~Roth, ``Lightweight probabilistic deep networks,'' in
  \emph{Proceedings of the IEEE Conference on Computer Vision and Pattern
  Recognition}, 2018, pp. 3369--3378.

\bibitem{r23}
A.~Kendall and Y.~Gal, ``What uncertainties do we need in bayesian deep
  learning for computer vision?'' \emph{Advances in neural information
  processing systems}, vol.~30, 2017.

\bibitem{r24}
H.~O. Hartley, ``The modified gauss-newton method for the fitting of non-linear
  regression functions by least squares,'' \emph{Technometrics}, vol.~3, no.~2,
  pp. 269--280, 1961.

\bibitem{r25}
A.~Geiger, P.~Lenz, and R.~Urtasun, ``Are we ready for autonomous driving? the
  kitti vision benchmark suite,'' in \emph{2012 IEEE conference on computer
  vision and pattern recognition}.\hskip 1em plus 0.5em minus 0.4em\relax IEEE,
  2012, pp. 3354--3361.

\bibitem{r26}
Y.~Liao, J.~Xie, and A.~Geiger, ``Kitti-360: A novel dataset and benchmarks for
  urban scene understanding in 2d and 3d,'' \emph{arXiv preprint
  arXiv:2109.13410}, 2021.

\bibitem{r27}
I.~Loshchilov and F.~Hutter, ``Fixing weight decay regularization in adam,''
  2018.

\bibitem{r28}
L.~N. Smith and N.~Topin, ``Super-convergence: Very fast training of neural
  networks using large learning rates,'' in \emph{Artificial intelligence and
  machine learning for multi-domain operations applications}, vol. 11006.\hskip
  1em plus 0.5em minus 0.4em\relax International Society for Optics and
  Photonics, 2019, p. 1100612.

\bibitem{r29}
K.~He, X.~Zhang, S.~Ren, and J.~Sun, ``Delving deep into rectifiers: Surpassing
  human-level performance on imagenet classification,'' in \emph{Proceedings of
  the IEEE international conference on computer vision}, 2015, pp. 1026--1034.

\end{thebibliography}

\end{document}